\documentclass[10pt,twocolumn,letterpaper]{article}

\usepackage[pagenumbers]{cvpr} 
\usepackage{graphicx}
\usepackage{booktabs}
\usepackage[most]{tcolorbox}
\usepackage{epsfig}
\usepackage{tabularx}
\usepackage{booktabs}
\usepackage{relsize}
\usepackage{multirow}
\usepackage{scrextend}           
\usepackage{array}               
\usepackage{makecell}            
\usepackage{wasysym}             
\usepackage{xspace}
\usepackage{amssymb} 
\usepackage{pifont} 
\usepackage{bm}
\usepackage{colortbl}
\usepackage[flushleft]{threeparttable}
\usepackage{subcaption}
\usepackage{pifont}
\usepackage{colortbl}
\usepackage{bm}
\usepackage{graphicx}
\usepackage{setspace}
\usepackage{cuted}
\usepackage{soul}
\usepackage{dblfloatfix}
\usepackage{esvect}
\usepackage{caption}
\definecolor{lightgray}{RGB}{235,235,235}
\definecolor{car}{RGB}{230,25,75}
\definecolor{cyclist}{RGB}{255, 130, 48}
\definecolor{pedestrian}{RGB}{138, 43, 226}
\definecolor{colorgt}{RGB}{170,255,195}
\definecolor{white}{rgb}{1.0, 1.0, 1.0}
\definecolor{monocop}{RGB}{71, 159, 179}
\definecolor{baseline}{RGB}{253, 190, 110}
\definecolor{gray}{RGB}{243, 243, 243}
\definecolor{darkGreen}{rgb}{0.01, 0.8, 0.24}

\newcommand{\monoThreeD}{Mono3D\xspace}

\newcommand{\twoD}{$2$D\xspace}
\newcommand{\threeD}{$3$D\xspace}

\newcommand{\iou}{IoU\xspace}

\newcommand{\iouThreeD}{IoU$_{3\text{D}}$\xspace}

\newcommand{\bev}{BEV\xspace}

\newcommand{\mathDash}{$-$}

\newcommand{\kitti}{KITTI\xspace}
\newcommand{\nuscenes}{nuScenes\xspace}
\newcommand{\waymo}{Waymo\xspace}

\newcommand{\valOne}{Val\xspace}

\newcommand{\val}{Val\xspace}
\newcommand{\train}{Train\xspace}
\newcommand{\test}{Test\xspace}
\newcommand{\frontal}{frontal\xspace}

\newcommand{\ap}{AP}

\newcommand{\apThreeD}{\ap$_{3\text{D}}$\xspace}
\newcommand{\aphThreeD}{APH$_{3\text{D}}$\xspace}

\newcommand{\apBev}{\ap$_{\text{BEV}}$\xspace}

\newcommand{\bracketPercentage}{[\%]}

\newcommand{\apThreeDSeventy}{\ap$_{\!3\text{D}\!}$ 70\xspace}

\newcommand{\sota}{SoTA\xspace}

\newcommand{\deviant}{DEVIANT\xspace}

\newcommand{\gupNet}{GUP Net\xspace}

\newcommand{\monodetr}{MonoDETR\xspace}
\newcommand{\monodgp}{MonoDGP\xspace}

\newcommand{\occupancymThreeD}{OccupancyM3D\xspace}
\newcommand{\opaThreeD}{OPA-3D\xspace}

\newcommand{\monocon}{MonoCon\xspace}
\newcommand{\fdThreeD}{FD3D\xspace}
\newcommand{\monoflex}{MonoFlex\xspace}
\newcommand{\monorcnn}{MonoRCNN\xspace}
\newcommand{\monouni}{MonoUNI\xspace}
\newcommand{\monocd}{MonoCD\xspace}
\newcommand{\monomae}{MonoMAE\xspace}
\newcommand{\monotakd}{MonoTAKD\xspace}

\providecommand\rightarrowRHD{\relbar\joinrel\mathrel\RHD}

\newcommand{\uparrowRHD}  {\rotatebox[origin=c]{90}{$\rightarrowRHD$}}

\newcommand{\uparrowRHDSmall}  {\raisebox{0.05\normalbaselineskip}{\scalebox{0.7}{\uparrowRHD}}}   

\usepackage{pifont}
\newcommand{\cmark}{\ding{51}}

\definecolor{XLcolor}{rgb}{0.858, 0.188, 0.478}

\definecolor{cvprblue}{rgb}{0.21,0.49,0.74}
\usepackage[pagebackref,breaklinks,colorlinks,allcolors=cvprblue]{hyperref}

\usepackage[capitalize]{cleveref}
\crefname{section}{Sec.}{Secs.}
\Crefname{section}{Section}{Sections}
\Crefname{table}{Table}{Tables}
\crefname{table}{Tab.}{Tabs.}

\newcommand{\methodName}{MonoCoP\xspace}
\newcommand{\methodNameFull}{Unleashing the Power of Chain-of-Prediction for Monocular \threeD Object Detection}

\newcommand{\chaindepthshort}{$z$}

\newcommand{\chainangleshort}{$\theta$}

\newcommand{\chaindimentionshort}{$\mathbf{s}$}

\newcommand{\queryenhancement}{AttributeNet\xspace}
\newcommand{\queryenhancementshort}{AN\xspace}
\def\paperID{} 
\def\confName{CVPR}
\def\confYear{2026}

\title{\methodNameFull}

\author{%
  Zhihao Zhang\textsuperscript{1}\quad Abhinav Kumar\textsuperscript{1} \quad Girish Chandar Ganesan\textsuperscript{1} \quad Xiaoming Liu\textsuperscript{1,2} \\
\textsuperscript{1}Michigan State University \quad \textsuperscript{2}University of North Carolina at Chapel Hill\\
  \{zhan2365, ganesang\}@msu.edu \quad abhinav3663@gmail.com\quad liuxm@cs.unc.edu
}

\begin{document}
\maketitle

\begin{strip}
  \centering
  \vspace{-6mm}
  \includegraphics[width= 1 \linewidth]{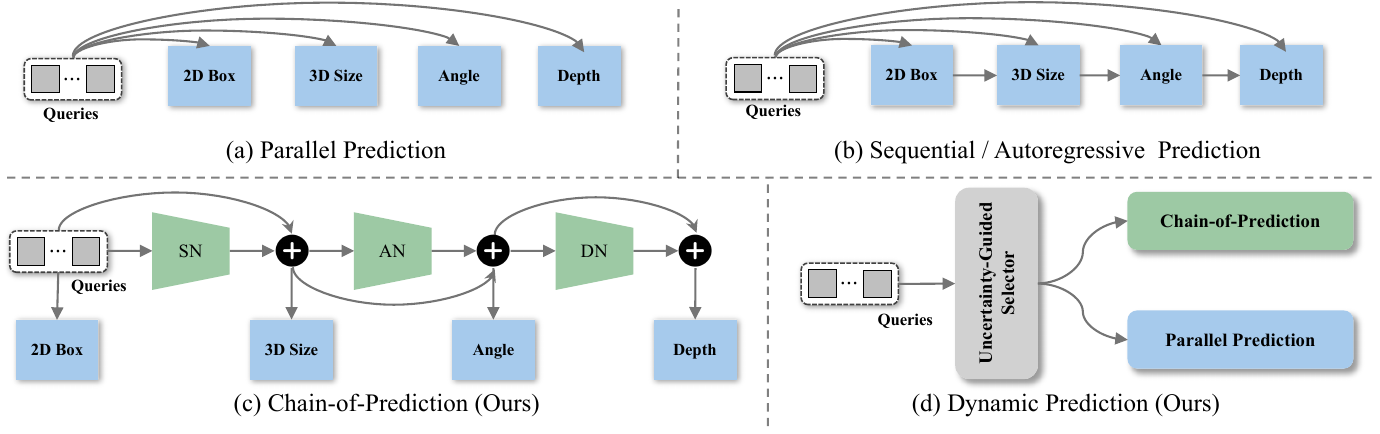}
  \vspace{-7mm}
    \captionof{figure}{
    Overview of prediction paradigms in \monoThreeD.
    (a) \textbf{Parallel Prediction}: predicts multiple \threeD attributes (\eg, size, orientation, depth) independently, ignoring their inter-dependencies.  
    (b) \textbf{Sequential Prediction}: predicts attributes step by step, conditioning each on the previously estimated ones, which easily causes error accumulation across attributes.
   (c) \textbf{Chain-of-Prediction (Ours)}: captures feature-level inter-attribute correlations by \textit{progressively learning, propagating, and aggregating attribute-specific features}, effectively mitigating error accumulation in b.
    (d) \textbf{Dynamic Prediction (Ours)}: \textit{dynamically switches} between CoP and parallel prediction for each object based on the predicted uncertainty, effectively combining strengths from both prediction paradigms.
    }

  \label{fig:intro-teaser}
\end{strip}

\begin{abstract}
Monocular \threeD detection (\monoThreeD) aims to infer \threeD bounding boxes from a single RGB image.
Without auxiliary sensors such as LiDAR, this task is inherently ill-posed since the \threeD-to-\twoD projection introduces depth ambiguity.
Previous works often predict \threeD attributes (\eg, depth, size, and orientation) in parallel, overlooking that these attributes are inherently correlated through the \threeD-to-\twoD projection.
However, simply enforcing such correlations through sequential prediction can propagate errors across attributes, especially when objects are occluded or truncated, where inaccurate size or orientation predictions can further amplify depth errors.
Therefore, neither parallel nor sequential prediction is optimal.
In this paper, we propose \methodName, an adaptive framework that learns when and how to leverage inter-attribute correlations with two complementary designs.
A Chain-of-Prediction (CoP) explores inter-attribute correlations through feature-level learning, propagation, and aggregation, while an Uncertainty-Guided Selector (GS) dynamically switches between CoP and parallel paradigms for each object based on the predicted uncertainty.
By combining their strengths, \methodName achieves state-of-the-art    performance on \kitti, \nuscenes, and \waymo, significantly improving depth accuracy, particularly for distant objects. Code and models are publicly available at  \href{https://github.com/alanzhangcs/MonoCoP}{https://github.com/alanzhangcs/MonoCoP}. 
\end{abstract}    
\vspace{-4mm}
\section{Introduction}
\label{sec:intro}

Monocular \threeD object detection (\monoThreeD) aims to infer an object’s \threeD properties (\eg, size, orientation, and depth) from a single RGB image. 
Compared with approaches that rely on LiDAR~\cite{yin2021center, shi2019pointrcnn, peng2024learning} or stereo cameras~\cite{li2019stereo, shi2022stereo}, \monoThreeD has attracted considerable attention for its cost efficiency, ease of deployment, and suitability for applications such as autonomous driving~\cite{simonelli2020disentangling} and robotics~\cite{ma20233d, zhang2024tamm，zhang2025physrig}.

However, without auxiliary depth sensors, \monoThreeD faces a fundamental challenge of ill-posed depth estimation~\cite{liu2020smoke, ma2021delving, lu2021geometry, kumar2025charm3r, yang2025monoclue}, which stems from the inherent ambiguity of recovering \threeD structure from a single \twoD image. 
To mitigate this issue, recent works have sought to extract richer depth cues from images. 
For instance, \monorcnn~\cite{MonoRCNN_ICCV21} estimates depth through geometric projection using \twoD box heights and known \threeD object dimensions, while \gupNet~\cite{lu2021geometry} models depth uncertainty to improve reliability. 
\monodetr~\cite{zhang2023monodetr} leverages object-wise depth supervision to inject explicit geometric priors. \monocd~\cite{yan2024monocd} provides complementary depth and \monodgp~\cite{pu2024monodgp} models depth error distributions to further refine predictions.

\begin{figure}[t]
    \centering
    \includegraphics[width=1\linewidth]{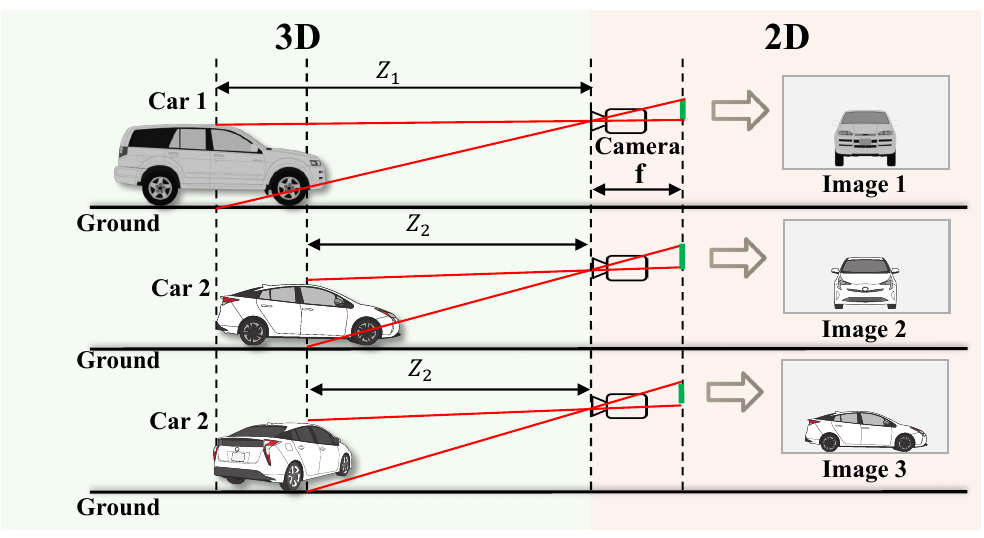}
    \vspace{-6mm}
    \caption{
        \textbf{Illustration of inter-correlated \threeD attributes} in \monoThreeD.
        Through the \threeD-to-\twoD projection, attributes such as depth, size, and orientation jointly determine an object’s appearance in the image, making them inherently coupled. 
        As shown in Images 1–2, cars at different depths appear with similar \twoD sizes when their \threeD sizes differ, while in Images 2–3, the same car at a fixed depth exhibits apparent scale changes under different orientations. 
        This projection-induced coupling leads to inherent ambiguity when inferring \threeD structure from a single \twoD image, highlighting the need to explicitly model their inter-correlations.
    }

    \label{fig:teaser_vis}
    \vspace{-6mm}
\end{figure}

Despite these improvements, an important observation is overlooked by the aforementioned methods:
\textit{depth, size, and orientation are inherently correlated through the \threeD-to-\twoD projection.}
During projection, these attributes jointly determine an object’s \twoD appearance, meaning that multiple \threeD configurations yield nearly identical visual observations.
As illustrated in \cref{fig:teaser_vis}, a nearby small car and a distant large car occupy almost the same \twoD region, while a single car viewed from different orientations exhibits varying apparent scales.
This projection-induced coupling makes it inherently ambiguous to infer one attribute (\eg, depth) without considering the others (\eg, size and orientation), emphasizing the necessity of modeling their inter-correlations.

A simple way to model such inter-attribute correlations is through sequential prediction, as adopted in prior works~\cite{liu2022autoregressive, xue2022point2seq, pang2025randar},  
where each \threeD attribute is predicted conditioned on the previously estimated ones (see~\cref{fig:intro-teaser}b).  
However, this conventional sequential strategy tends to amplify estimation errors,  
as inaccuracies in one attribute are likely to propagate to others.  
Moreover, for objects whose attributes are inherently uncertain or difficult to estimate,  
such dependencies further magnify these errors, leading to degraded overall performance.  
\textit{Thus, neither parallel prediction nor sequential prediction yields an optimal solution.}

To address the limitations of both parallel and sequential prediction, we propose \methodName, an adaptive framework that learns \textit{when and how to leverage inter-attribute correlations} through two complementary designs.  
First, instead of predicting \threeD attributes step by step as in prior sequential approaches~\cite{liu2022autoregressive, brazil2019pedestrian,xue2022point2seq, pang2025randar} (see~\cref{fig:intro-teaser}b),  
\methodName introduces a \textbf{Chain-of-Prediction (CoP)} paradigm that models inter-attribute correlations directly at the feature level.  
CoP explicitly learns, propagates, and aggregates attribute-specific features (see~\cref{fig:intro-teaser}c), effectively reducing the error accumulation inherent in conventional sequential prediction through joint feature-level optimization.
Second, we design an \textbf{Uncertainty-Guided Selector (GS)} that assesses the depth uncertainty of both CoP and parallel branches for each object and dynamically selects the more reliable one (see~\cref{fig:intro-teaser}d), effectively combining the strengths of both paradigms.  
Together, CoP and GS  balance correlation exploitation and independence preservation across diverse objects, enabling adaptive and robust \monoThreeD{}.

In summary, our main contributions are as follows:
\begin{itemize}[leftmargin=*, noitemsep, nolistsep]

\item We mathematically illustrate that depth, size, and orientation are correlated through the \threeD-to-\twoD projection.

\item We observe that the benefit of modeling inter-attribute correlations varies across objects, making both purely parallel and purely sequential prediction suboptimal.

\item We introduce \methodName, an adaptive framework that learns when and how to leverage inter-attribute correlations through (1) a Chain-of-Prediction (CoP) that models feature-level correlations within a single forward pass, and (2) an Uncertainty-Guided Selector (GS) that dynamically selects between chain and parallel prediction.

\item Extensive experiments on \kitti, \nuscenes, and \waymo demonstrate \methodName achieves state-of-the-art~(\sota) performance, delivering consistent gains in both near and distant object detection.

\end{itemize}

\section{Related Work}
\label{sec:relatted}
\noindent\textbf{\monoThreeD .} There are two lines of work based on architectural differences. 1) CNN~\cite{ren2016faster} based \monoThreeD  methods~\cite{brazil2019m3d, kumar2021groomed, zhang2021objects, liu2020smoke, ma2021delving}. Some focus on center-based pipelines~\cite{liu2020smoke, ma2021delving, wang2022probabilistic, liu2023monocular}. Some exploit geometric relations between \twoD and \threeD~\cite{li2022diversity, lu2021geometry, zhang2021objects, wu2024fd3d, brazil2020kinematic} to improve the accuracy of \threeD detection, while others use depth-equivariant blocks~\cite{kumar2022deviant, qin2022monoground}, or adopt \twoD detector FPN~\cite{brazil2019m3d, kumar2021groomed, brazil2023omni3d}.  Some also incorporate extra training data, such as \threeD CAD models~\cite{chabot2017deep, liu2021autoshape, lee2023baam}, LiDAR point clouds~\cite{ma2019accurate, wang2019pseudo, chen2021monorun, reading2021categorical, peng2024learning, chong2022monodistill, gui2024remote, huang2024training, liu2025monotakd, long2025riccardo, long2023radiant}, synthetic data~\cite{lin2025drivegen, zhu2023tame}, BEV encoding~\cite{kumar2024seabird} and dense depth maps~\cite{ding2020learning, qin2021monogrnet, jiang2024weakly, ma2020rethinking, peng2022did, ranasinghe2024monodiff, zhu2024replay} which enable models to implicitly learn depth features during training.  We refer to~\cite{ma20233d} for this survey.
2) Transformer~\cite{carion2020end, chen2023group, zhao2024detrs, kim2025monodino} based \monoThreeD  methods~\cite{huang2022monodtr, wu2023monopgc, zhou2023monoatt, zhang2023monodetr, pu2024monodgp}. These methods introduce visual transformers~\cite{carion2020end, zhu2010deformable} to \threeD detectors without NMS or anchors, achieving higher accuracies. For example, \monodetr ~\cite{zhang2023monodetr} introduces a unique depth-guided transformer to improve \threeD detection with depth-enhanced queries. \monodgp~\cite{pu2024monodgp}  further develops a decoupled visual decoder with error-based depth estimation. However, these methods overlook the inter-correlations among \threeD attributes when inferring them from \twoD images. 
\methodName explicitly models these correlations and adaptively selects between correlated and independent predictions, leading to more robust and accurate \threeD detection.

\noindent\textbf{Depth Estimation in \monoThreeD.} 
Depth estimation remains the key bottleneck in \monoThreeD~\cite{ma2021delving, li2022diversity, lu2021geometry, yan2024monocd, pu2024monodgp}. 
Recent works improve depth prediction by introducing geometric priors~\cite{lu2021geometry}, multi-hypothesis modeling~\cite{li2022diversity}, multi-branch fusion~\cite{zhang2023monodetr}, and uncertainty calibration~\cite{pu2024monodgp, yan2024monocd}. 
However, these methods estimate depth in isolation, ignoring the inter-correlations among \threeD attributes. 
In contrast, our \methodName jointly models these inter-correlations at the feature level and dynamically switches between CoP and parallel prediction for different objects.

\noindent\textbf{Sequential and Autoregressive Prediction.} 
Sequential and autoregressive paradigms are widely used in LLMs~\cite{liu2023llava, wei2022chain}, image~\cite{tian2024visual, sun2024autoregressive, wu2025janus, zhang2023tile}, and video generation~\cite{deng2024autoregressive, pang2025randar, xu2026emotag} to capture structured dependencies by conditioning each step on prior outputs. 
This idea has also been explored in point cloud \threeD detection~\cite{liu2022autoregressive, xue2022point2seq}, where \threeD attributes are estimated sequentially to model interrelations. 
However, these approaches operate on prediction outputs rather than latent features, making them prone to cumulative errors and preventing joint optimization across attributes. 
We extend this paradigm to the \emph{feature level}, where it explicitly learns, propagates, and aggregates attribute specific features.

\section{Inter-Correlations of \threeD Attributes}
\label{sec:interconrrelations}
\noindent\textbf{\monoThreeD Definition.}
\monoThreeD takes a single RGB image together with camera parameters as input and aims to localize and classify objects in \threeD metric space.
Each object is represented by its category $C$, a \twoD bounding box $B_{2D}$ on the image plane, and a \threeD bounding box $B_{3D}$ in real-world coordinates.
The 3D box $B_{3D}$ is parameterized by its center $\mathbf{c}=(x_c,y_c,z_c)$, \threeD size $\mathbf{s}=(w,h,l)$, and orientation $\theta$, all defined in the camera-centered metric coordinate system.

\subsection{Empirical Observation}
To empirically examine whether the predicted \threeD attributes exhibit statistical dependencies, 
we analyze predictions on the \kitti \val set using a trained \monodetr detector~\cite{zhang2023monodetr}. 
True positives with $\mathrm{IoU_{2D}}\ge0.7$ are retained, and Pearson correlation coefficients are computed between the per-attribute prediction errors of depth, size, and orientation. 
Orientation errors are wrapped to $(-\pi,\pi]$ using $\mathrm{atan2}(\sin\Delta\theta,\cos\Delta\theta)$ for angular consistency. 
We observe that depth errors have a weak-to-moderate correlation ($r=0.35$) with size errors and a weak but consistent positive correlation with orientation errors ($r=0.11$). 
Although these correlations are modest in magnitude, they reveal that the predicted \threeD attributes are not fully independent, providing empirical support that depth estimation benefits from modeling its relationships with other correlated attributes.

\subsection{Analytical Evidence}
The statistical dependencies observed above are explained analytically by the geometry of the \threeD-to-\twoD projection. 
Under the pinhole camera model, a \threeD point $\mathbf{p}=(X,Y,Z)$ projects to:
\begin{equation}
u = \frac{fX}{Z} + c_x,\qquad v = \frac{fY}{Z} + c_y,
\end{equation}
where $f$ denotes the focal length and $(c_x,c_y)$ the principal point.
Consider a box corner with local offset $\mathbf{\Delta}=[\pm l/2,\pm w/2,\pm h/2]^T$.
After rotation $\mathbf{R}(\theta)$ and translation $\mathbf{c}$, its camera-frame coordinate becomes
$\mathbf{p}=\mathbf{c}+\mathbf{R}(\theta)\mathbf{\Delta}$.
Define:
\begin{equation}
\alpha(\mathbf{s},\theta)=[\mathbf{R}(\theta)\mathbf{\Delta}]_x,\qquad
\beta(\mathbf{s},\theta)=[\mathbf{R}(\theta)\mathbf{\Delta}]_z,
\end{equation}
so that the horizontal projection is expressed as:
\begin{equation}
F(\mathbf{s},\theta,z_c)=\frac{f(x_c+\alpha)}{z_c+\beta}+c_x.
\end{equation}
For a fixed observed projection $u = u_0$, we analyze how the object orientation $\theta$ influences the corresponding depth $z_c$ under the projection constraint $F(\mathbf{s}, \theta, z_c) = u_0$. 
By differentiating this relation with respect to $\theta$ (while keeping $\mathbf{s}$ constant), we obtain:
\begin{equation}
\frac{dz_c}{d\theta}
=\frac{\alpha'(z_c+\beta)-(x_c+\alpha)\beta'}{x_c+\alpha},
\end{equation}
where $\alpha'=\partial\alpha/\partial\theta$ and $\beta'=\partial\beta/\partial\theta$.
Except for degenerate cases, $\frac{dz_c}{d\theta}\neq0$, indicating that orientation $\theta$ and depth $z_c$ are inherently coupled through the projection geometry.
A similar derivation along the vertical axis shows that the projected height depends jointly on object size and depth, revealing that depth and size are also coupled.
Therefore, the \threeD attributes $\{z_c, \mathbf{s}, \theta\}$ are geometrically inter-dependent rather than separable, motivating models that explicitly capture such inter-attribute coupling.

\begin{figure*}[t]
    \centering
    \includegraphics[width=1\linewidth]{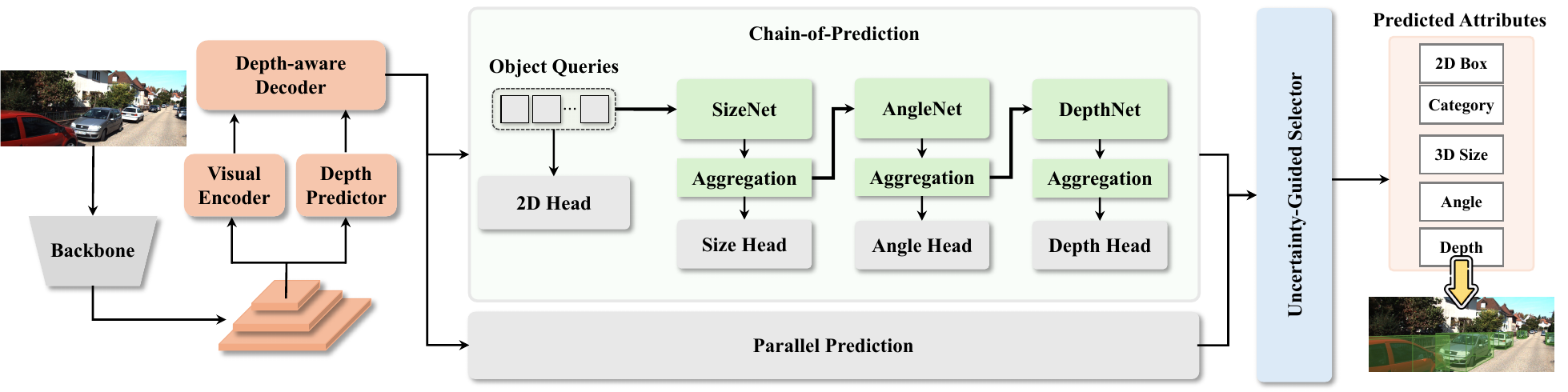}
    \caption{\textbf{\methodName\ Overview.}
    \threeD attributes (\eg, depth, size, and orientation) are \textbf{correlated} through the \threeD-to-\twoD projection. 
    \methodName\ learns \textit{when} and \textit{how} to exploit these correlations through two complementary modules.
    The \textbf{Chain-of-Prediction (CoP)} captures cross-attribute dependencies at the feature level, progressively propagating and aggregating attribute-specific cues to enhance geometric consistency and mitigate error accumulation. 
    The \textbf{Uncertainty-Guided Selector (GS)} adaptively selects between CoP and parallel pathways for each object based on its depth uncertainty, combining their strengths to achieve more accurate and robust \threeD detection.
    }

    \label{sec:method-model-arch1}
    \vspace{-3mm}
\end{figure*}

\section{\methodName }
\label{sec:method}

\noindent\textbf{Overview.}
We propose \methodName, a unified framework that adaptively models inter-attribute correlations in \monoThreeD.  
The key observation is that depth, size, and orientation are geometrically coupled through the \threeD-to-\twoD projection. However, the benefit of modeling inter-attribute correlations varies across objects, making both purely parallel and purely sequential prediction suboptimal. To address this variability, \methodName integrates two complementary designs.
The \textit{Chain-of-Prediction (CoP)} explicitly captures inter-attribute dependencies at the feature level, progressively propagating and aggregating attribute-specific features within a single forward pass.  
Meanwhile, the \textit{Uncertainty-Guided Selector (GS)} monitors prediction uncertainty for each object and dynamically selects between the chain-based and parallel pathways.  
By leveraging inter-attribute correlations and bypassing uncertain dependencies, \methodName achieves more accurate \threeD detection.

\subsection{Chain-of-Prediction}

The Chain-of-Prediction (CoP) module serves as the core of \methodName, designed to capture inter-dependencies among correlated \threeD attributes.  
Instead of predicting all attributes in parallel, the model processes them sequentially through three stages: 
\textit{Feature Learning}, \textit{Feature Propagation}, and \textit{Feature Aggregation}.  

\noindent\textbf{Feature Learning.}  
To enable feature-level sequential prediction, we first learn attribute-specific features for each \threeD attribute.  
A lightweight \queryenhancement (\queryenhancementshort) module is applied to the object query $\mathbf{q}$ 
to obtain three features corresponding to \threeD size, orientation, and depth:
\begin{equation}
\mathbf{f}_s = A_s(\mathbf{q}), \quad
\mathbf{f}_a = A_a(\mathbf{q}), \quad
\mathbf{f}_d = A_d(\mathbf{q}),
\end{equation}
where each submodule $A(\cdot)$ consists of two linear layers with a nonlinear activation:
\begin{equation}
A(\mathbf{q}) = \sigma(\mathbf{qW_1})\mathbf{W_2}.
\end{equation}
These attribute-specific features form the basis for learning structured dependencies among \threeD attributes.

\begin{figure*}[t]
    \centering
    \begin{subfigure}[b]{0.32\textwidth}
        \centering
        \includegraphics[width=\textwidth]{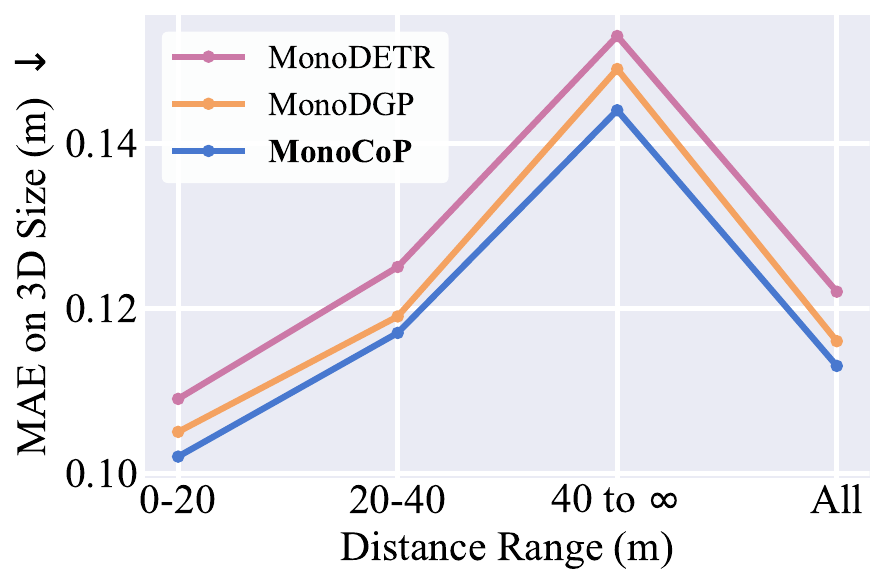} 
        \caption{\threeD Size Error.}
        \label{fig:sub1}
    \end{subfigure}
    \hfill
    \begin{subfigure}[b]{0.32\textwidth}
        \centering
        \includegraphics[width=\textwidth]{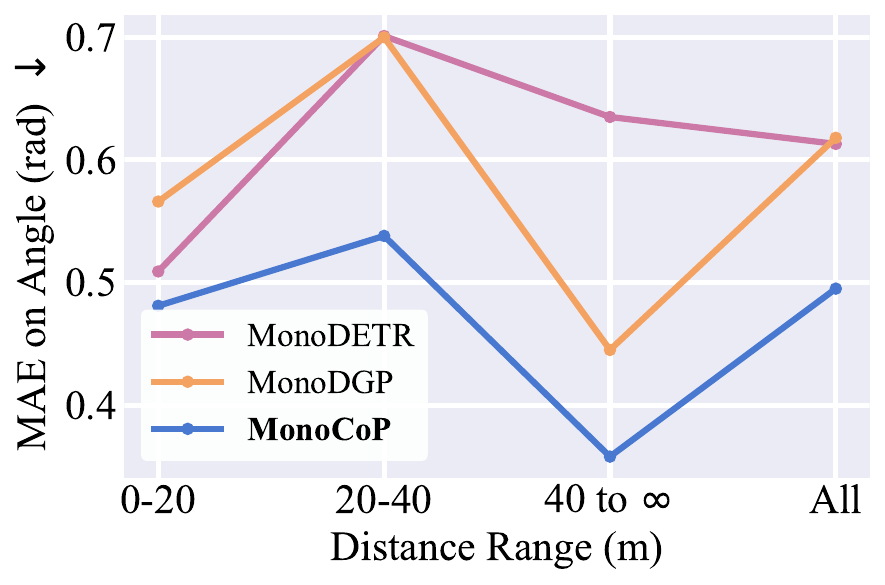} 
        \caption{Orientation Error.}
        \label{fig:sub2}
    \end{subfigure}
    \hfill
    \begin{subfigure}[b]{0.32\textwidth}
        \centering
        \includegraphics[width=\textwidth]{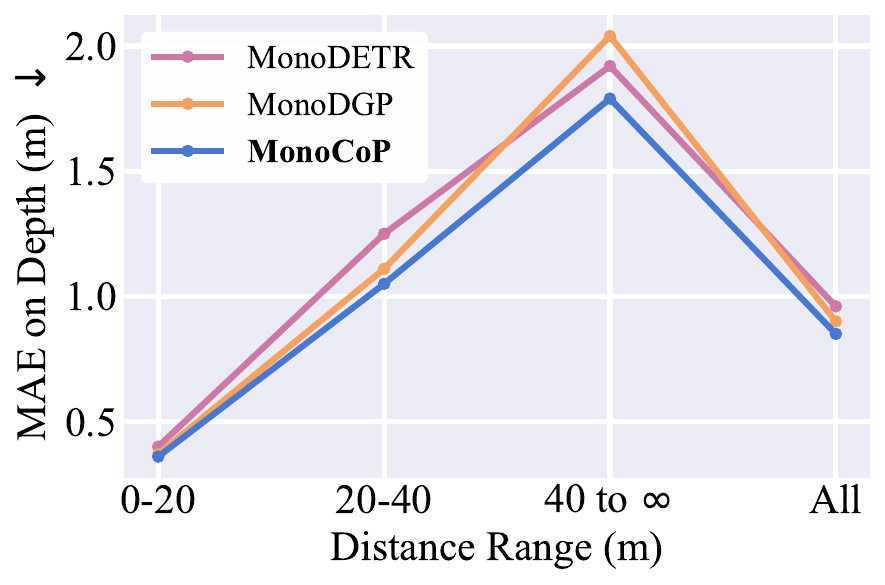} 
        \caption{Depth Error.}
        \label{fig:sub3}
    \end{subfigure}
    \vspace{-2mm}
    \caption{\textbf{Mean Absolute Error (MAE) on \kitti \val.} We compute the MAE for predicted \threeD attributes (3D size, orientation, and depth) across multiple distance ranges. Compared to previous parallel prediction approaches~\cite{zhang2023monodetr, pu2024monodgp}, \methodName consistently yields lower errors, particularly for distant objects,  demonstrating that our \methodName outperforms conventional parallel prediction strategies.}
    \label{fig:error}
    \vspace{-4mm} 
    
\end{figure*}

\noindent\textbf{Feature Propagation.}  
Although attribute-specific features are obtained, they remain independent of one another.  
To capture their inter-dependencies, \methodName constructs a sequential chain, where the feature learned for one attribute guides the prediction of the next.  
This stepwise propagation allows later predictions to benefit from earlier cues.  
The prediction order follows a progression from \threeD size to orientation and finally to depth, as these attributes demand increasing levels of spatial understanding:  
dimension prediction primarily focuses on object extent, orientation requires reasoning about \threeD rotation, and depth estimation necessitates full spatial understanding.  
Formally, the chain is defined as:
\begin{equation}
\mathbf{f}_s = A_s(\mathbf{q}), \quad
\mathbf{f}_a = A_a(\mathbf{f}_s), \quad
\mathbf{f}_d = A_d(\mathbf{f}_a),
\end{equation}
enabling a progressive flow of information from size to orientation and finally to depth.

\noindent\textbf{Feature Aggregation.}  
Purely sequential propagation leads to feature forgetting and error accumulation along the chain.  
To address this, we incorporate residual aggregation~\cite{he2016deep, veit2016residual} so that each stage preserves information from all previous ones.  
At each step, the predicted feature is combined with its input to form an aggregated representation:
\begin{equation}
\tilde{\mathbf{f}}_s = A_s(\mathbf{q}) + \mathbf{q}, \hspace{0.5em}
\tilde{\mathbf{f}}_a = A_a(\tilde{\mathbf{f}}_s) + \tilde{\mathbf{f}}_s, \hspace{0.5em} 
\tilde{\mathbf{f}}_d = A_d(\tilde{\mathbf{f}}_a) + \tilde{\mathbf{f}}_a.
\end{equation}
This residual aggregation ensures that each attribute prediction benefits from all preceding features, mitigating feature forgetting and improving overall depth stability.

\subsection{Uncertainty-Guided Selector}

The CoP strengthens inter-attribute correlation modeling by sequentially propagating and aggregating attribute-specific features across prediction stages.
However, the reliability of these learned dependencies varies across objects.
When objects are partially occluded or lack clear visual cues, the predicted orientation and \threeD size become unreliable, which in turn makes the final depth estimation unstable and allows errors in one attribute to propagate to others.
Therefore, a fixed sequential dependency is not always optimal.
To address this issue, we introduce an Uncertainty-Guided Selector (GS), which monitors object-level depth uncertainty and dynamically switches between CoP and parallel pathways.

\noindent\textbf{Uncertainty Estimation.}  
Following probabilistic depth modeling~\cite{lu2021geometry}, we assume that the predicted depth $\hat{z}$ follows a Laplace distribution centered at the ground-truth depth $z^*$ with scale parameter $\sigma$, representing the \textit{aleatoric uncertainty} of the prediction:
\begin{equation}
p(z^*|\hat{z}, \sigma) = \frac{1}{2\sigma}\exp\!\left(-\frac{|z^* - \hat{z}|}{\sigma}\right).
\end{equation}
Minimizing the negative log-likelihood yields the following depth loss:
\begin{equation}
\mathcal{L}_{\text{depth}} =
\sqrt{2}\, e^{-\log\sigma} \left| \hat{z} - z^* \right| + \log\sigma,
\label{eq:uncertainty_loss}
\end{equation}
which enables the model to jointly predict both the expected depth and its corresponding uncertainty.  
The predicted $\sigma$ thus reflects the confidence level of depth estimation and serves as a continuous signal for object-level reliability.

\noindent\textbf{Selecting Mechanism.}  
During training, the GS associates each object’s reliability with the inverse of its predicted uncertainty, defined as \( r = 1 / \sigma \).  
For each query, the chain-based pathway produces a reliability score \( \tilde{r}^{(\text{CoP})} \).  
We introduce a threshold hyper-parameter \( \tau \) to determine whether the chain-based reasoning is sufficiently reliable.  
Specifically, when the reliability falls below the threshold, the router switches to the parallel pathway; otherwise, the chain-based pathway is used:
\begin{equation}
b^* =
\begin{cases}
\text{CoP}, & \text{if } \tilde{r}^{(\text{CoP})} \ge \tau, \\
\text{Par}, & \text{otherwise}.
\end{cases}
\end{equation}
\textit{Intuitively, GS follows the chain-based pathway when attribute correlations can be estimated with sufficient confidence, and switches to the parallel pathway when the predicted uncertainty indicates unreliable prediction.}

\begin{table*}[t]
    \centering
    \tabcolsep=0.1cm
    \resizebox{0.99\textwidth}{!}{
        \begin{tabular}{l|c|c|ccc|ccc|ccc|ccc}
            \toprule
            \multirow{2}{*}{Method} & \multirow{2}{*} {\makecell{Extra \\ Data}} & \multirow{2}{*}{Venue} 
            & \multicolumn{3}{c|}{ \textbf{\val }, \apThreeD (\uparrowRHDSmall) } 
            & \multicolumn{3}{c|}{ \textbf{\val }, \apBev (\uparrowRHDSmall) } 
            & \multicolumn{3}{c|}{ \textbf{\test }, \apThreeD (\uparrowRHDSmall) } 
            & \multicolumn{3}{c}{ \textbf{\test }, \apBev (\uparrowRHDSmall) }\\
            
            & & & Easy & Mod. & Hard & Easy & Mod. & Hard & Easy & Mod. & Hard & Easy & Mod. & Hard\\ 
            \midrule
            \occupancymThreeD~\cite{peng2024learning} & LiDAR & CVPR~24  
            & 26.87 & 19.96 & 17.15 & 35.72 & 26.60 & 23.68 & 25.55 & 17.02 & 14.79 & 35.38 & 24.18 & 21.37\\
            \opaThreeD~\cite{su2023opa} & Depth & ICRA~23 
            & 24.97 & 19.40 & 16.59 & 33.80 & 25.51 & 22.13 & 24.68 & 17.17 & 14.14 & 32.50 & 23.14 & 20.30\\
            \monotakd~\cite{liu2025monotakd}* & LiDAR & CVPR 25 & 34.36 & 22.61 & 19.88 & 42.86 & 29.41 & 26.47 & 27.91 & 19.43 & 16.51 & 38.75 & 27.76 & 24.14\\
            \midrule
            \monoflex~\cite{zhang2021objects} & \multirow{12}{*}{None} & CVPR~21 
            & 23.64 & 17.51 & 14.83 & \mathDash  & \mathDash  & \mathDash  & 19.94 & 13.89 & 12.07 & 28.23 & 19.75 & 16.89 \\
            \monorcnn~\cite{MonoRCNN_ICCV21} &  & ICCV~21 
            & \mathDash  & \mathDash  & \mathDash  &  \mathDash  & \mathDash  & \mathDash   & 18.36 & 12.65 & 10.03 & 25.48 & 18.11 & 14.10 \\
            \gupNet~\cite{lu2021geometry}  &  & CVPR~21 
            & 22.76 & 16.46 & 13.72 & 31.07 & 22.94 & 19.75 & 20.11 & 14.20 & 11.77 & \mathDash  & \mathDash  & \mathDash  \\
            \deviant~\cite{kumar2022deviant}  &  & ECCV 22 
            & 24.63 & 16.54 & 14.52 & 32.60 & 23.04 & 19.99 & 21.88 & 14.46 & 11.89 & 29.65 & 20.44 & 17.43\\
            \monocon{}~\cite{liu2022monocon}  &  & AAAI 22 
            & 26.33 & 19.01 & 15.98 & \mathDash  & \mathDash  & \mathDash  & 22.50 & 16.46 & 13.95 & 31.12 & 22.10 & 19.00 \\
            \monouni~\cite{jia2023monouni} &  & NeurIPS 23 
            & 24.51 & 17.18 & 14.01 & \mathDash  & \mathDash  & \mathDash  & 24.75 & 16.73 & 13.49 & \mathDash  & \mathDash  & \mathDash \\
           
            \monodetr~\cite{zhang2023monodetr} &  & ICCV 23 
            & 28.84 & 20.61 & 16.38 & 37.86 & 26.95 & 22.80 & 25.00 & 16.47 & 13.58 & 33.60 & 22.11 & 18.60 \\
            \monocd~\cite{yan2024monocd} &  & CVPR~24 
            & 26.45 & 19.37 & 16.38 & 34.60 & 24.96 & 21.51 & 25.53 & 16.59 & 14.53 & 33.41 & 22.81 & 19.57 \\
            \fdThreeD~\cite{wu2024fd3d} &  & AAAI~24 
            & 28.22 & 20.23 & 17.04 & 36.98 & 26.77 & 23.16 & 25.38 & 17.12 & 14.50 & 34.20 & 23.72 & 20.76 \\
            \monomae~\cite{jiang2024monomae} &  & NeurIPS 24 
            & 30.29 & 20.90 & 17.61 & 40.26 & 27.08 &  23.14 & 25.60 & \underline{18.84} & \textbf{16.78} & 34.15 & 24.93 & 21.76 \\
            \monodgp~\cite{pu2024monodgp} &  & CVPR 25  
            & \underline{30.76} & \underline{22.34} & \underline{19.02} & \underline{39.40} & \underline{28.20} & \underline{24.42} & \underline{26.35} & 18.72 & 15.97 & \underline{35.24} & \underline{25.23} & \underline{22.02} \\
            \rowcolor{lightgray}\textbf{\methodName  (Ours)} &  & CVPR 26   
            & \textbf{32.06} & \textbf{23.98} & \textbf{20.64} & \textbf{42.20} & \textbf{31.29} & \textbf{27.58} & \textbf{27.54} & \textbf{19.11} & \underline{16.33} & \textbf{36.77} & \textbf{25.57} & \textbf{22.62} \\

            \bottomrule
        \end{tabular}}
        \caption{
        \textbf{\kitti \val and \test results} at \iouThreeD $\ge 0.7$.  
        Under the setting without using any extra data, \methodName\ achieves \sota performance across most metrics, 
        surpassing all RGB-only counterparts by notable margins and performing comparably to methods that leverage LiDAR or depth supervision.
        [Key: \textbf{First}, \underline{Second}, *=Knowledge Distillation]
        }
    \vspace{-3mm}
    \label{exp:test-car}
\end{table*}

\begin{table}[t]
    \centering
    \resizebox{0.48 \textwidth}{!}{
    \begin{tabular}{l|c|cc|cc}
        \toprule
        \multirow{2}{*}{Method} & \multirow{2}{*}{\iou    } & \multicolumn{2}{c|}{\apThreeD{}} & \multicolumn{2}{c}{\apBev{}} \\

         & &  Easy & Mod. & Easy & Mod. \\
        \midrule
        \gupNet~\cite{lu2021geometry} & \multirow{5}{*}{0.7} & 8.50 & 7.40 & 14.21 & 12.81 \\
        \deviant~\cite{kumar2022deviant} & & 9.69 & 8.33 & 16.28 & 14.36 \\
        \monodetr~\cite{zhang2023monodetr} &  & 9.53 & 8.19 & 16.39 & 14.41 \\
        \monodgp~\cite{pu2024monodgp} & & \underline{10.04} &  \underline{8.78} & \underline{16.55} & \underline{14.53} \\ 
        \rowcolor{lightgray} \textbf{\methodName  (Ours)} & & \textbf{10.85} & \textbf{9.71} & \textbf{17.83} & \textbf{15.86}  \\
        \midrule
        \gupNet~\cite{lu2021geometry} & \multirow{5}{*}{0.5} & 29.03 & 26.16 & 33.42 & 30.23 \\
        \deviant~\cite{kumar2022deviant} & & 31.47 & 28.22 & 35.61 & 31.93 \\
        \monodetr~\cite{zhang2023monodetr} &  & \underline{31.81} & \underline{28.35} & \underline{35.70} & \underline{31.96} \\
        \monodgp~\cite{pu2024monodgp} & &  29.56 & 26.17 & 32.67 & 29.44\\ 
        \rowcolor{lightgray} \textbf{\methodName (Ours)} & & \textbf{33.70} & \textbf{29.91} & \textbf{37.44} & \textbf{34.01}  \\
        \bottomrule
    \end{tabular}
    }
    \caption{
        \textbf{\nuscenes \val Results.} \methodName  achieves \sota on \threeD and \bev detection across two thresholds. [Key: \textbf{First}, \underline{Second}]}
    \label{sec:experiment-nuscenes} 
    \vspace{-4mm}
\end{table}

{
\setlength\tabcolsep{2pt}
\begin{table*}[t]
    \centering
    \resizebox{1 \textwidth}{!}{
    \begin{tabular}{l|c|cccc|cccc|cccc|cccc}
        \toprule
        \multirow{3}{*}{Method} & \multirow{3}{*}{Difficulty} & \multicolumn{8}{c|}{\iouThreeD     $\geq$ 0.5} & \multicolumn{8}{c}{\iouThreeD     $\geq$ 0.7} \\
        
        & & \multicolumn{4}{c|}{\aphThreeD   \bracketPercentage (\uparrowRHDSmall)} & \multicolumn{4}{c|}{\apThreeD{} \bracketPercentage (\uparrowRHDSmall)} & \multicolumn{4}{c|}{\aphThreeD   \bracketPercentage (\uparrowRHDSmall)} & \multicolumn{4}{c}{\apThreeD{} \bracketPercentage (\uparrowRHDSmall)}\\ 
        
        &  & All & 0-30 & 30-50 & 50-$\infty$ & All & 0-30 & 30-50 & 50-$\infty$ & All & 0-30 & 30-50 & 50-$\infty$ & All & 0-30 & 30-50 & 50-$\infty$ \\
        \midrule
        \gupNet~\cite{lu2021geometry} in \cite{kumar2022deviant}   & \multirow{6}{*}{Level 1}  
        & 9.94 & 24.59 & 4.78 & 0.22 & 10.02 & 24.78 & 4.84 & 0.22 & 2.27 & 6.11 & 0.80 & 0.03 & 2.28 & 6.15 & 0.81 & 0.03 \\
        \deviant~\cite{kumar2022deviant} & & \underline{10.89} & \underline{26.64} & \underline{5.08} & 0.18 & \underline{10.98} & \underline{26.85} & \underline{5.13} & 0.18 & 2.67 & 6.90 & \underline{0.98} & 0.02 & 2.69 & 6.95 & \underline{0.99} & 0.02 \\
        \monodetr~\cite{zhang2023monodetr}\textsuperscript{$\dagger$} & & 9.60 & 23.58 & 4.67 & \underline{0.99} & 9.68 & 23.78 & 4.72 & \underline{1.00} & 2.10 & 5.94 & 0.73 & \underline{0.12} & 2.11 & 5.99 & 0.73 & 0.12 \\
        \monouni~\cite{jia2023monouni}   & & 10.73 & 26.30 & 3.98 & 0.55 & \underline{10.98} & 26.63 & 4.04 & 0.57 & \textbf{3.16} & \textbf{8.50} & 0.86 & \underline{0.12} & \textbf{3.20} & \textbf{8.61} & 0.87 & \underline{0.13} \\
        \monodgp~\cite{pu2024monodgp}\textsuperscript{$\dagger$} & & 9.84 & 23.73 & 5.01 & 0.98 & 10.06 & 24.01 & 5.06 & 0.99 & 2.39 & 6.62 & 0.84 & \underline{0.12} & 2.41 & 6.67 & 0.84 & 0.12 \\
        \rowcolor{lightgray} \textbf{\methodName  (Ours)} & & \textbf{11.65} & \textbf{27.35} & \textbf{5.97} & \textbf{1.46} & \textbf{11.76} & \textbf{27.59} & \textbf{6.03} & \textbf{1.48} & \underline{2.70} & \underline{7.38} & \textbf{1.06} & \textbf{0.16} & \underline{2.72} & \underline{7.44} & \textbf{1.07} & \textbf{0.16} \\
        \midrule
        \gupNet~\cite{lu2021geometry} in \cite{kumar2022deviant}   &  \multirow{6}{*}{Level 2} 
        & 9.31 & 24.50 & 4.62 & 0.19 & 9.39 & 24.69 & 4.67 & 0.19 & 2.12 & 6.08 & 0.77 & 0.02 & 2.14 & 6.13 & 0.78 & 0.02 \\
        \deviant~\cite{kumar2022deviant} &  & 10.20 & \underline{26.54} & \underline{4.90} & 0.16 & 10.29 & \underline{26.75} & \underline{4.95} & 0.16 & 2.50 & 6.87 & \underline{0.94} & 0.02 & 2.52 & 6.93 & \underline{0.95} & 0.02 \\
        \monodetr~\cite{zhang2023monodetr}\textsuperscript{$\dagger$} &  & 9.00 & 23.49 & 4.51 & \underline{0.86} & 9.08 & 23.70 & 4.55 & \underline{0.87} & 1.97 & 5.92 & 0.70 & 0.10 & 1.98 & 5.96 & 0.71 & 0.10 \\
        \monouni~\cite{jia2023monouni}   &  & \underline{10.24} & 26.24 & 3.89 & 0.51 & \underline{10.38} & 26.57 & 3.95 & 0.53 & \textbf{3.00} & \textbf{8.48} & 0.84 & \underline{0.12} & \textbf{3.04} & \textbf{8.59} & 0.85 & \underline{0.12} \\
        \monodgp~\cite{pu2024monodgp}\textsuperscript{$\dagger$} &  & 9.32 & 23.65 & 4.84 & 0.85 & 9.43 & 23.92 & 4.88 & 0.86 & 2.24 & 6.59 & 0.81 & 0.10 & 2.26 & 6.65 & 0.81 & 0.10 \\
        \rowcolor{lightgray} \textbf{\methodName  (Ours)} &  & \textbf{10.93} & \textbf{27.25} & \textbf{5.76} & \textbf{1.27} & \textbf{11.03} & \textbf{27.49} & \textbf{5.82} & \textbf{1.29} & \underline{2.53} & \underline{7.35} & \textbf{1.02} & \textbf{0.14} & \underline{2.55} & \underline{7.41} & \textbf{1.03} & \textbf{0.14} \\
        \bottomrule
        \end{tabular}}
    \caption{\textbf{\waymo \valOne Vehicle results.} 
    \methodName consistently outperforms all methods on most metrics across both difficulty levels (Level $1$ and Level $2$) and IoU thresholds ($0.5$ and $0.7$). 
    [Key: \textbf{First}, \underline{Second}, $\dagger$= Retrained]}
    \label{exp:mono3d_waymo}
    \vspace{-2mm}
\end{table*}
}

\begin{table}[t]
    \centering
    \tabcolsep=0.05cm
    \resizebox{0.4\textwidth}{!}{
    \begin{tabular}{l|c|ccc}
    \toprule
         Method &  \apThreeDSeventy (\uparrowRHDSmall) & \#Param (M) & GFLOPs  \\
         \midrule
         
         \monodetr~\cite{zhang2023monodetr} &  21.12 & 35.93 & 62.12 \\
         \monodgp~\cite{pu2024monodgp}  & 22.34 & 38.90 & 68.99 \\
         \textbf{\methodName (Ours)} &  23.98 & 42.50 & 71.77 \\
        \bottomrule
    \end{tabular}}
    \vspace{-2mm}
    \caption{\textbf{Comparison of efficiency metrics.} 
    \methodName attains higher accuracy with only marginal increases in parameters and computation, achieving a superior accuracy–efficiency trade-off.}
    \label{exp:param}
    \vspace{-6mm}
\end{table}

\section{Experiments}
\label{sec:experiments}

\subsection{Experimental Settings}
\noindent \textbf{Datasets.} We evaluate our method on three datasets.

\noindent$\bullet$ \kitti  ~\cite{geiger2012we} is a widely used benchmark with $7{,}481$ training images and $7{,}518$ testing images. It includes three classes: Car, Pedestrian, and Cyclist.  All objects are divided into three difficulty levels: Easy, Moderate, and Hard. \cite{chen2016monocular} further partitions the $7{,}481$ training samples of \kitti into $3{,}712$ training and $3{,}769$ validation images.

\noindent$\bullet$ \waymo \cite{sun2020scalability} is a large-scale \threeD dataset. Following~\cite{kumar2022deviant}, we use images from the front camera, splitting the dataset into $52{,}386$ training images and $39{,}848$ validation images. \waymo defines two object levels: Level~$1$ and Level~$2$. Each object is assigned a level based on the number of LiDAR points contained within its \threeD bounding box.

\noindent $\bullet$ \nuscenes~\cite{caesar2020nuscenes} has $28,130$ training images and $6,019$ validation images captured by front camera. We follow~\cite{kumar2022deviant} to transform its labels into \kitti style. As \nuscenes does not provide truncation or occlusion labels,  objects are categorized into two groups based on \twoD height: Easy, Moderate.

\noindent\textbf{Evaluation Metrics.} For KITTI, we use the \apThreeD{} and \apBev{} metrics with IoU thresholds of $0.7$ for Cars and $0.5$ for Pedestrians and Cyclists, respectively, in the Moderate category to benchmark models~\cite{simonelli2019disentangling}.  We also use the mean absolute error (MAE) between predicted \threeD attributes and ground truth attributes. For \waymo, we use the \aphThreeD metric, which incorporates heading information, to benchmark models, following~\cite{reading2021categorical, kumar2022deviant}. Additionally, we report results at three distance ranges: $[0{,}30)$, $[30{,}50)$, and $[50{,}\infty)$ meters. For \nuscenes, we adopt the \kitti metrics. 

\noindent\textbf{Implementation Details.} 
We follow \monodetr~\cite{zhang2023monodetr} and \monodgp~\cite{pu2024monodgp}, and conduct all experiments on a single NVIDIA A6000 GPU. Following the training protocol of \monodgp~\cite{pu2024monodgp}, we train \methodName from scratch for 250 epochs using an initial learning rate of $2 \times 10^{-4}$. Additional implementation details are provided in Appendix~\ref{appx-imple}.

\subsection{Main Results} 
\noindent\textbf{\kitti \val Results.}   
\cref{exp:test-car}  presents detection results on the \kitti \val set.  
We report the median performance of the best checkpoint across five independent runs for fair comparison.  
\methodName achieves \sota \threeD detection accuracy, surpassing all existing methods.  
The largest gains are observed in the \textit{Moderate} set, with substantial improvements also in the \textit{Easy} and \textit{Hard} sets.  
Remarkably, \methodName even outperforms methods trained with additional data, demonstrating both efficiency and robustness. 
\cref{fig:error} additionally reports the MAE between predicted and ground-truth \threeD attributes.  
Compared with prior works~\cite{zhang2023monodetr, pu2024monodgp} that adopt parallel prediction, \methodName consistently achieves lower errors across all distance ranges, confirming its superiority in modeling inter-attribute dependencies.  
Notably, the improvement becomes more pronounced for distant objects, where depth ambiguity is more severe, further validating the effectiveness of our approach.

\noindent \textbf{\kitti\ Leaderboard (\test) Results.}  
As shown in \cref{exp:test-car}, \methodName\ consistently achieves \sota performance across all metrics for both \threeD and \bev detection on the \kitti\ leaderboard.   For fair comparison, we explicitly specify the use of additional data in our report.  
When trained \emph{without} any extra data, \methodName\ surpasses prior methods by $1.19$ AP under the \textit{Easy} setting for \threeD detection.  
Even when compared against approaches trained with auxiliary data, \methodName\ maintains superior performance across all metrics, underscoring its strong generalization capability.

{
\setlength\tabcolsep{2.5pt}
\begin{table*}[t]
\centering
\begin{subtable}[t]{0.25\textwidth}
\centering
\tabcolsep=0.18cm
\resizebox{\textwidth}{!}{%
\begin{tabular}{@{}cc|ccc}
\multirow{2}{*}{CoP} & \multirow{2}{*}{GS}  & \multicolumn{3}{c}{\val, \apThreeD (\uparrowRHDSmall)}   \\
&   & Easy & Mod. & Hard \\
\midrule
& & 29.41 & 21.12 & 18.11 \\
\cmark & &  \textbf{32.40} & 23.64 & 20.31 \\
\rowcolor{lightgray} \cmark & \cmark & 32.06 &  \textbf{23.98} & \textbf{20.64} \\
\end{tabular}}
\caption{
\textbf{Components of \methodName.}
Both proposed CoP and GS contribute to improvements.
}
\label{sec:experiment-ablation:components}
\end{subtable}
\hfill
\begin{subtable}[t]{0.31\textwidth}
\centering
\tabcolsep=0.18cm
\resizebox{\textwidth}{!}{%
\begin{tabular}{@{}c|c|ccc}
\multirow{2}{*}{Routers} 
& \multirow{2}{*}{Acc(\%)} 
& \multicolumn{3}{c}{\val, \apThreeD (\uparrowRHDSmall)}  \\
& & Easy & Mod. & Hard \\
\midrule
w/gt & 100.00 & 32.95 & 24.11 & 21.55  \\
\midrule
Random & 50.00 & 30.84 & 22.35 & 18.56  \\
\rowcolor{lightgray} GS & \cellcolor{lightgray}\textbf{82.18} & \cellcolor{lightgray}\textbf{32.06} & \cellcolor{lightgray}\textbf{23.98} & \cellcolor{lightgray}\textbf{20.64} \\
\end{tabular}}
\caption{\textbf{Router design.} GS  dynamically switches between CoP and parallel pathways, yielding near-oracle performance in all difficulty levels.}

\label{sec:experiment-ablation:router-design}
\end{subtable}
\hfill
\begin{subtable}[t]{0.26\textwidth}
\centering
\tabcolsep=0.18cm
\resizebox{\textwidth}{!}{%
\begin{tabular}{@{}c|ccc}
\multirow{2}{*}{Alternatives} & \multicolumn{3}{c}{\val, \apThreeD (\uparrowRHDSmall)} \\
& Easy & Mod. & Hard  \\
\midrule
HTL~\cite{lu2021geometry} & 25.15 & 18.42 & 15.52 \\
CoOp~\cite{zhou2022learning} & 28.83 & 21.23 & 17.76 \\
\cellcolor{lightgray}\methodName & \cellcolor{lightgray}\textbf{32.06} & \cellcolor{lightgray}\textbf{23.98} & \cellcolor{lightgray}\textbf{20.64}  \\
\end{tabular}}
\caption{\textbf{Alternatives.} \methodName effectively models attribute correlations  compared to other alternatives.}
\label{sec:experiment-ablation:alternatives}
\end{subtable}

\begin{subtable}[t]{0.26\textwidth}
\centering
\tabcolsep=0.18cm
\resizebox{\textwidth}{!}{%
\begin{tabular}{@{}ccc|ccc}
\multirow{2}{*}{FL} & \multirow{2}{*}{FP} & \multirow{2}{*}{FA}  & \multicolumn{3}{c}{\val, \apThreeD (\uparrowRHDSmall)}  \\
& & & Easy & Mod. & Hard \\
\midrule
& & & 29.41 & 21.12 & 18.11 \\
\cmark & & & 29.67 & 21.74 & 18.23  \\
\cmark & \cmark &  & 29.33 & 22.22 & 19.26 \\
\rowcolor{lightgray} \cmark & \cmark & \cmark & \textbf{32.06} &  \textbf{23.98} & \textbf{20.64} \\

\end{tabular}}
\caption{
\textbf{CoP design.}
Feature Learning (FL), Propagation (FP), and Aggregation (FA) in CoP progressively strengthen dependency modeling across attributes.
}
\label{sec:experiment-ablation:cop-design}
\end{subtable}
\hfill
\begin{subtable}[t]{0.25\textwidth}
\centering
\tabcolsep=0.18cm
\resizebox{\textwidth}{!}{%
\begin{tabular}{@{}c|ccc}
\multirow{2}{*}{\makecell{Prediction \\ Order}} & \multicolumn{3}{c}{\val, \apThreeD (\uparrowRHDSmall)}  \\
& Easy & Mod. & Hard \\
\midrule
\chaindepthshort $\vv{}$ \chaindimentionshort $\vv{}$ \chainangleshort & 30.54 & 22.54 & 19.37 \\
\chainangleshort $\vv{}$ \chaindimentionshort $\vv{}$ \chaindepthshort & 29.87 & 23.08 & 19.62\\
\rowcolor{lightgray} \chaindimentionshort $\vv{}$ \chainangleshort $\vv{}$ \chaindepthshort & \cellcolor{lightgray}\textbf{32.06} & \cellcolor{lightgray}\textbf{23.98} & \cellcolor{lightgray}\textbf{20.64} \\
\end{tabular}}
\caption{
\textbf{Prediction order in CoP.}
 Dimension (\chaindimentionshort), orientation (\chainangleshort), and then depth (\chaindepthshort) achieves the best,
aligning with geometric dependency among attributes.
}
\label{sec:experiment-ablation:order}
\end{subtable}
\hfill
\begin{subtable}[t]{0.27\textwidth}
\centering
\tabcolsep=0.18cm
\resizebox{\textwidth}{!}{%
\begin{tabular}{@{}c|ccc}
\multirow{2}{*}{\makecell{Backbone}} & \multicolumn{3}{c}{\val, \apThreeD (\uparrowRHDSmall)}  \\
& Easy & Mod. & Hard \\
\midrule
ResNet-34 & 28.32 & 22.32 & 19.23 \\
\rowcolor{lightgray} ResNet-50 & \cellcolor{lightgray}\textbf{32.06} & \cellcolor{lightgray}\textbf{23.98} & \cellcolor{lightgray}\textbf{20.64} \\
ResNet-101 & 30.14 & 21.75 & 18.56
\end{tabular}}
\caption{
\textbf{Backbone selection in \methodName.} Among all backbones, ResNet-50 provides the optimal balance between accuracy and efficiency for \methodName.
}
\label{sec:experiment-ablation:backbone}
\end{subtable}

\vspace{-2mm}
\caption{
\textbf{Ablation study of components in \methodName.}
We train \methodName from scratch on \kitti~\train\ and evaluate its \threeD detection performance on \kitti~\val\ under two \iou thresholds.
The configuration adopted by \methodName is \sethlcolor{lightgray}\hl{highlighted}.
}
\label{sec:experiment-ablation}
\vspace{-4mm}
\end{table*}
}

\noindent\textbf{\nuscenes \val  Results.}  \cref{sec:experiment-nuscenes}  shows results on \nuscenes \frontal dataset.  \methodName  significantly outperforms baselines across various \iou   thresholds and difficulty levels. Notably, the greatest improvements are observed on  the Mod.~set, particularly at \iouThreeD  $\geq 0.5$, underscoring the consistent performance gains achieved by our \methodName.

\noindent\textbf{\waymo\ \val\ Results.}  
To further evaluate the effectiveness of our method beyond \kitti, we conduct experiments on the large-scale \waymo\ dataset~\cite{sun2020scalability}, which presents greater scene diversity and scale variation.  
As shown in \cref{exp:mono3d_waymo}, \methodName\ significantly outperforms all image-only baselines at the $0.5$ \iou\ threshold, surpassing the previous best results by $0.78$ \aphThreeD\ and $0.76$ \apThreeD\ on Level~1 objects.  
At the stricter $0.7$ \iou\ threshold, \methodName\ remains highly competitive, ranking within the top two across all metrics.  
These results confirm that \methodName\ enhances both near- and far-range detection, demonstrating strong robustness and generalization.

\begin{figure*}[t]
  \centering
  \includegraphics[width= 1 \linewidth]{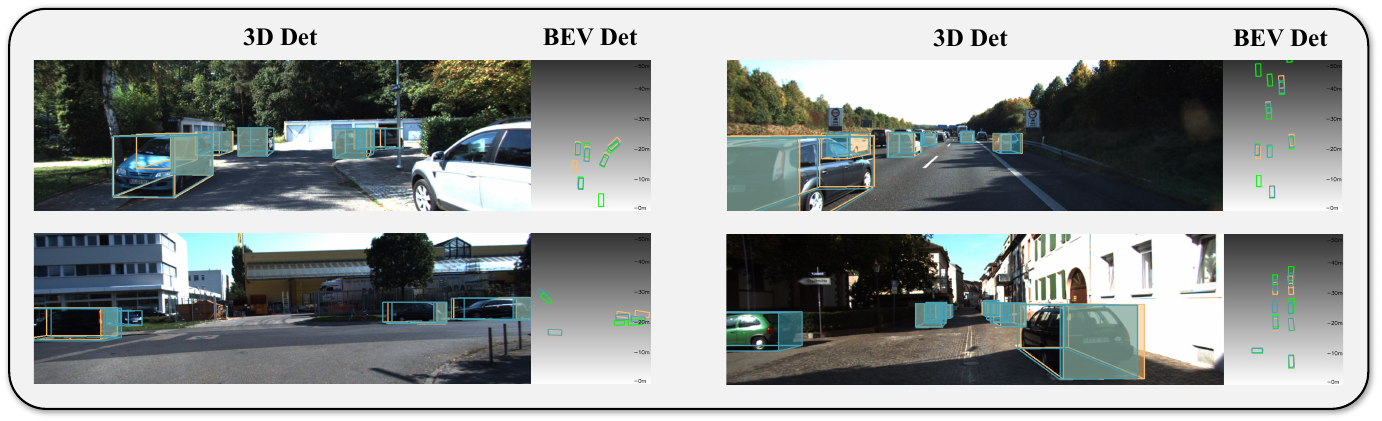}
  \vspace{-6mm}
  \caption{\textbf{Qualitative Results.} 
    \methodName improves detection accuracy, particularly for distant objects, consistent with the results in \cref{fig:error}c. 
  [Key: \textcolor{monocop}{\methodName}, \textcolor{baseline}{Baseline}, \textcolor{green}{Ground Truth}] }
  \label{sec:exp-visu}
  \vspace{-5mm}
\end{figure*}

\subsection{Efficiency Analysis.}
\cref{exp:param} summarizes the efficiency comparison.
\methodName adds only a $3.60M$ parameter overhead over current \sota \monodgp, yet yields a clear $+2.86$ gain in \apThreeD.
Thanks to the lightweight two-layer design of AttributeNet, computation remains nearly unchanged, with merely $+2.78$ GFLOPs, underscoring the superior efficiency–accuracy balance of our approach.

\noindent\textbf{Qualitative Results.} \cref{sec:exp-visu} visualizes the \threeD and BEV detection results on the \kitti \val set. We observe that \methodName improves detection accuracy, particularly for distant objects over the baseline~\cite{zhang2023monodetr}, consistent with the results in \cref{fig:error}c. We provide more visualizations on the \nuscenes \val and \waymo \val in Appendix~\ref{appx-vis}.

\subsection{Ablation Study}
We evaluate the individual design choices of \methodName on the \kitti \val split. Following standard practice, we treat the \apThreeD Mod. score at the 0.7 threshold as the primary evaluation metric.
We provide additional ablation studies in Appendix~\ref{appx-more-ablations}.

\noindent\textbf{Component.} 
\cref{sec:experiment-ablation:components} analyzes the impact of the proposed Chain-of-Prediction (CoP) and Uncertainty-Guided Selector (GS).
Starting from the baseline, introducing CoP yields consistent improvements across all IoU thresholds, confirming that explicitly modeling inter-attribute correlations enhances geometric consistency and depth accuracy.
When further equipped with UR, the model achieves an additional gain, reaching $+2.86\%$, with the largest improvements observed on the \textit{Moderate} and \textit{Hard} sets containing occluded or visually ambiguous objects.
A slight drop appears on the \textit{Easy} set, where most objects are clearly visible.
This observation is consistent with~\cref{sec:experiment-ablation:router-design}, as GS prioritizes reliability under uncertainty, but its routing decisions are not guaranteed to be correct in all cases.
Overall, the combined CoP and GS design delivers complementary benefits and consistent performance gains over the baseline.

\noindent\textbf{Router Design.}
To assess the effectiveness of the Uncertainty-Guided Selector (GS), we compare it with several routing strategies in~\cref{sec:experiment-ablation:router-design}.  
The \textit{w/gt} setting provides an oracle upper bound by selecting, for each object, the branch (CoP or Parallel) that yields the lower prediction error based on ground truth.  
In contrast, the \textit{Random} baseline randomly assigns each instance to either branch with equal probability, serving as a lower bound.  

As shown in the table~\cref{sec:experiment-ablation:router-design}, GS achieves an accuracy of $82.18\%$, approaching the oracle router while substantially surpassing the random strategy.  
This indicates that GS effectively learns to associate uncertainty with prediction reliability, enabling it to make consistent routing decisions.  
Correspondingly, GS attains significant performance gains across all difficulty levels, especially on the \textit{Hard} subset, where uncertainty estimation plays a more critical role under occlusion or ambiguous visual cues.  
Overall, GS demonstrates its ability to dynamically balance correlation exploitation and error mitigation, achieving accuracy close to the oracle performance without ground-truth supervision.

\noindent\textbf{CoP Design.}
Our Chain of Prediction (CoP) framework consists of three components: \textit{Feature Learning (FL)}, \textit{Feature Propagation (FP)}, and \textit{Feature Aggregation (FA)}. These components respectively learn, propagate, and aggregate attribute-specific features, enabling each attribute prediction to be conditioned on the preceding ones and effectively capture inter-attribute dependencies. Tab.~\ref{sec:experiment-ablation:cop-design} shows all three components contribute to performance improvements, and their combination achieves the best results.

\noindent\textbf{Prediction Order.}
\cref{sec:experiment-ablation:order} analyzes the effect of different prediction orders within the Chain-of-Prediction.  
The order of predicting dimension, orientation, and then depth achieves the best performance.  
This sequence follows a natural progression from \threeD size to orientation and finally to depth, as these attributes require progressively richer spatial cues:  
dimension prediction focuses on object extent, orientation depends on understanding \threeD rotation, and depth estimation benefits from comprehensive spatial context provided by the preceding attributes.  
Such dependency-aware ordering leads to more accurate \monoThreeD.

\noindent\textbf{Alternatives.}
We further investigate alternative approaches for modeling inter-correlations among \threeD attributes in~\cref{sec:experiment-ablation:alternatives}. We select two approaches: 1) HTL (Hierarchical Task Learning)~\cite{lu2021geometry} divides the training process into multiple stages, where each attribute is optimized sequentially; 2) CoOp~\cite{zhou2022learning} learns a learnable embedding for each attribute. Tab.~\ref{sec:experiment-ablation:alternatives} shows CoOp yields a slight improvement, whereas HTL causes a performance drop. In contrast, \methodName consistently outperforms both alternatives, further demonstrating the effectiveness of our method.

\noindent\textbf{Backbone Selection.}  \cref{sec:experiment-ablation:backbone} presents the detection performance of \methodName under different backbones and among them, ResNet-50 offers the best trade-off between accuracy and efficiency.


\section{Conclusion}
In this work, we explore the inter-correlations among \threeD attributes inferred from \twoD images and reveal that their benefits vary across objects. While parallel prediction neglects these geometric dependencies, rigid sequential prediction can propagate errors, making neither paradigm optimal. To address this challenge, we propose \methodName, an adaptive framework that learns when and how to exploit inter-attribute correlations. 
Extensive experiments show that \methodName consistently surpasses previous approaches and achieves \sota performance across multiple benchmarks including \kitti, \nuscenes and \waymo.

{
    \small
    \bibliographystyle{ieeenat_fullname}
    \bibliography{main}
}
\clearpage

\section*{\Large{Appendix}}
\setcounter{section}{0}
\setcounter{figure}{0}
\setcounter{table}{0}
\makeatletter 
\renewcommand{\thesection}{\Alph{section}}
\renewcommand{\theHsection}{\Alph{section}}
\renewcommand{\thefigure}{A\arabic{figure}}
\renewcommand{\theHfigure}{A\arabic{figure}}
\renewcommand{\thetable}{A\arabic{table}}
\renewcommand{\theHtable}{A\arabic{table}}
\makeatother
\renewcommand{\thetable}{A\arabic{table}}
\setcounter{equation}{0}
\renewcommand{\theequation}{A\arabic{equation}}

\section{Implementation Details}
\label{appx-imple}
In our implementation, we build \methodName upon the \monodetr framework~\cite{zhang2023monodetr}. All experiments are conducted on a single NVIDIA A6000 GPU. We train the model for 250 epochs using a batch size of 16 and a learning rate of $2 \times 10^{,-4}$. The AdamW optimizer is adopted with a weight decay of $10^{,-4}$. Additional hyperparameters and implementation details are provided in~\cref{tab:hyper}.

\begin{table}[!ht]
    \centering
    \renewcommand{\arraystretch}{1.15}
    \setlength{\tabcolsep}{15pt}
    \begin{tabular}{l|l}
        \toprule
        \textbf{Item} & \textbf{Value} \\
        \midrule
  
          optimizer & AdamW \\
          learning rate & 2e-4  \\
          weight decay & 1e-4 \\
          scheduler & Step \\ 
          decay rate & 0.5 \\
          decay list & [85, 125, 165, 205] \\
          
          number of feature scales & 4 \\
          hidden dim & 256 \\
          feedforward dim & 256 \\
          dropout & 0.1 \\
          nheads & 8 \\
          number of queries & 50 \\
          number of encoder layers & 3 \\
          number of decoder layers & 3 \\
          encoder npoints & 4 \\
          decoder npoints & 4 \\
          number of queries & 50 \\
          number of group & 11 \\
          class loss weight & 2 \\
          $\alpha$ in class loss & 0.25 \\ 
          bbox loss weight & 5 \\
          GIoU loss weight & 2 \\
          3D centor loss weight & 10 \\
          dim loss weight  & 1 \\
          depth loss weight & 1 \\
          depth map  loss weight & 1 \\
        
          class cost weight & 2 \\
          bbox cost weight & 5  \\
          GIoU cost weight & 2  \\ 
          3D centor cost weight & 10 \\ 
          \bottomrule
          \end{tabular}
    \caption{\textbf{Main hyperparameters of \methodName.}}
    \label{tab:hyper}
\end{table}

\section{Visualization}
\label{appx-vis}
\begin{figure*}[t]
    \centering
    \includegraphics[width= 0.95 \linewidth]{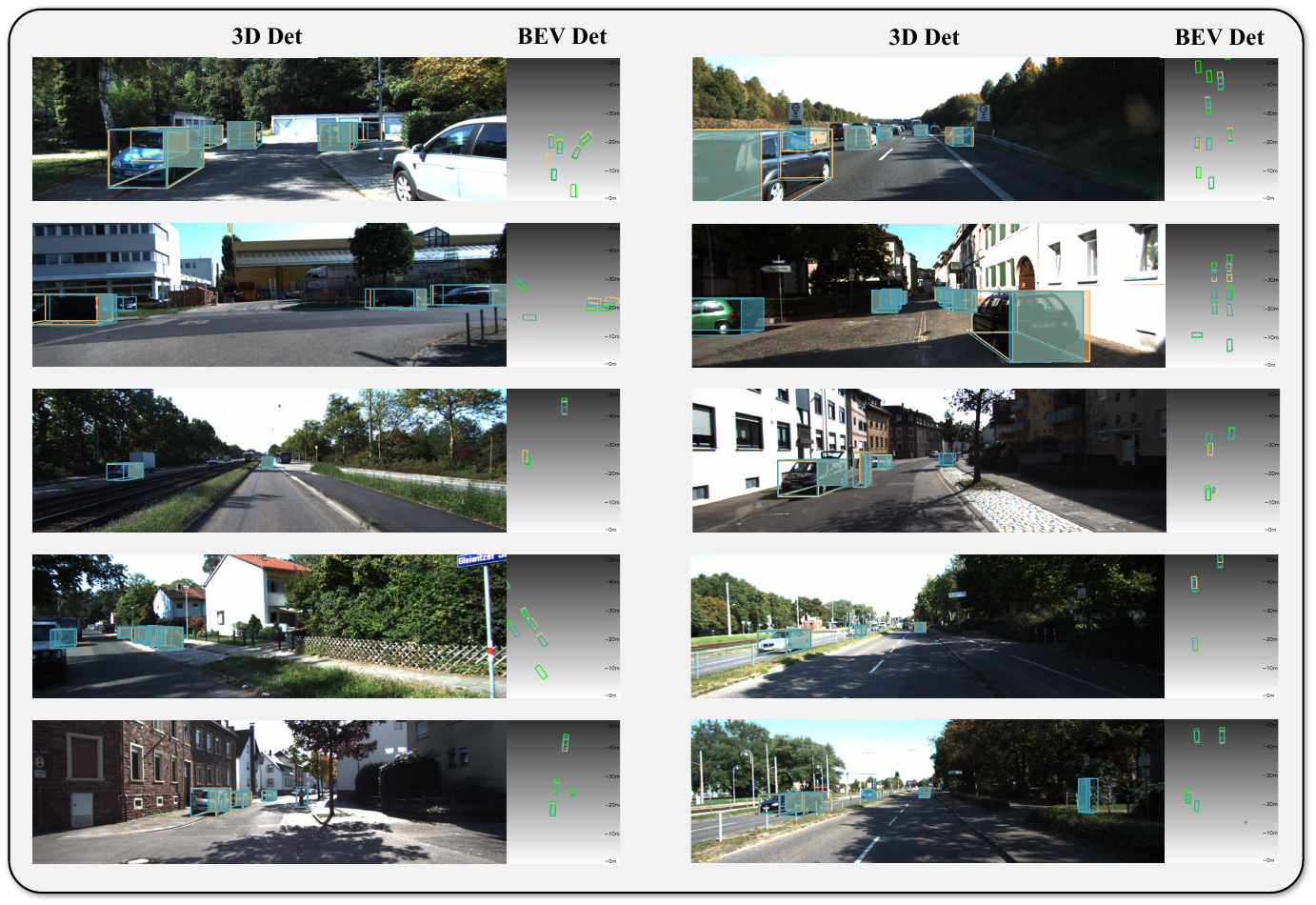}
    \caption{\textbf{\kitti Qualitative Results.} \methodName demonstrates superior performance in both \threeD and BEV detection over the baseline~\cite{zhang2023monodetr}. By predicting \threeD attributes conditionally to mitigate the instability and inaccuracy arising from their inter-correlation, \methodName improves detection accuracy, particularly for farther away objects, consistent with the results in Fig.~\ref{fig:error}c. [Key: \textcolor{monocop}{\methodName}, \textcolor{baseline}{Baseline}, \textcolor{green}{Ground Truth}] }
    \vspace{-4mm}
    \label{kitti-more}
\end{figure*}

\begin{figure*}[t]
    \centering
    \includegraphics[width= 0.95 \linewidth]{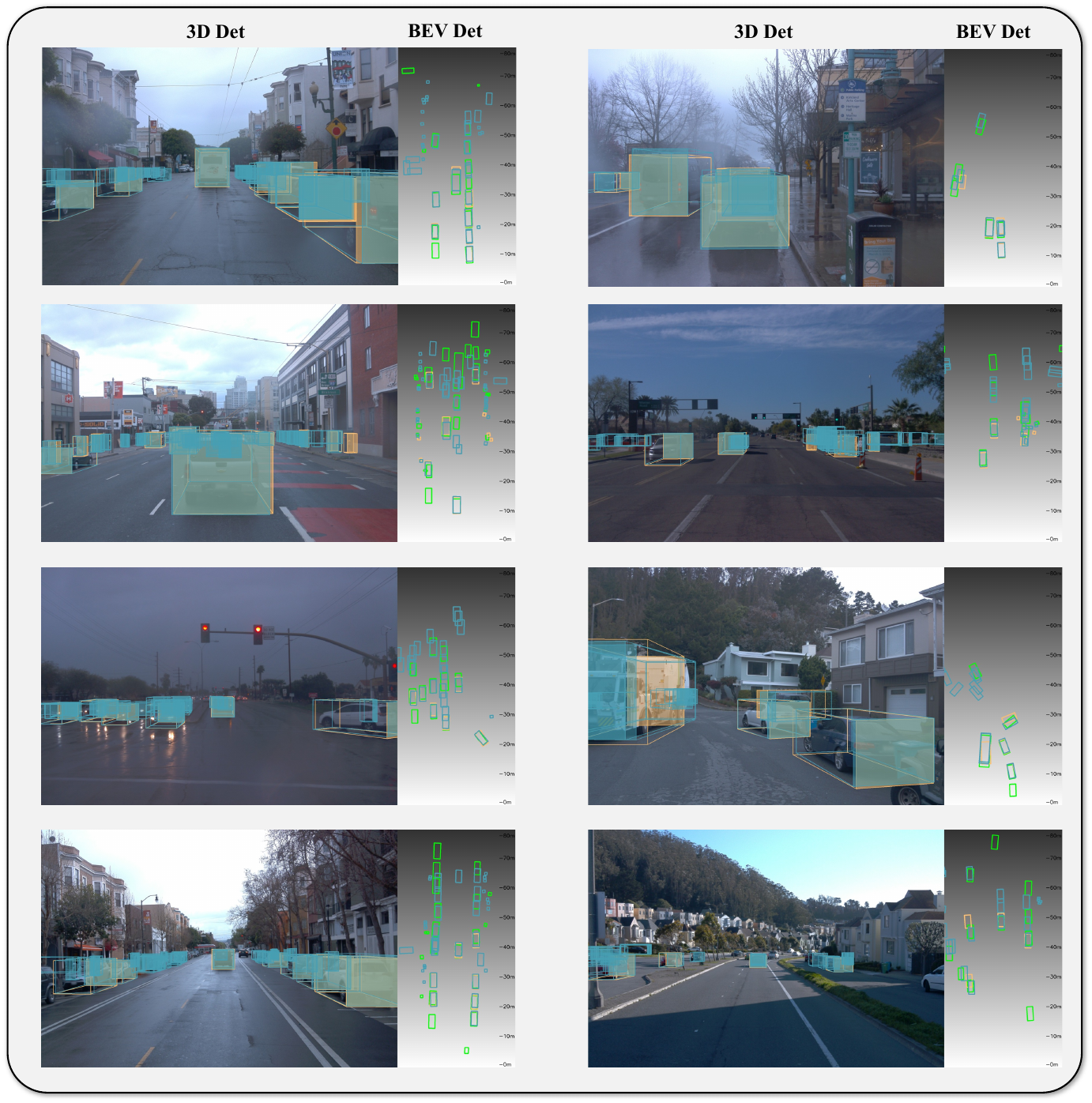}
    \caption{\textbf{\waymo Qualitative Results.} \methodName demonstrates superior performance in both \threeD and BEV detection over the baseline~\cite{zhang2023monodetr}. By predicting \threeD attributes conditionally to mitigate the instability and inaccuracy arising from their inter-correlations, \methodName predicts more accurate \threeD size and depth for large object, demonstrating the effectiveness of \methodName. [Key: \textcolor{monocop}{\methodName}, \textcolor{baseline}{Baseline}, \textcolor{green}{Ground Truth}] }
    \vspace{-4mm}
    \label{waymo-more}
\end{figure*}

\begin{figure*}[t]
    \centering
    \includegraphics[width= 0.95 \linewidth]{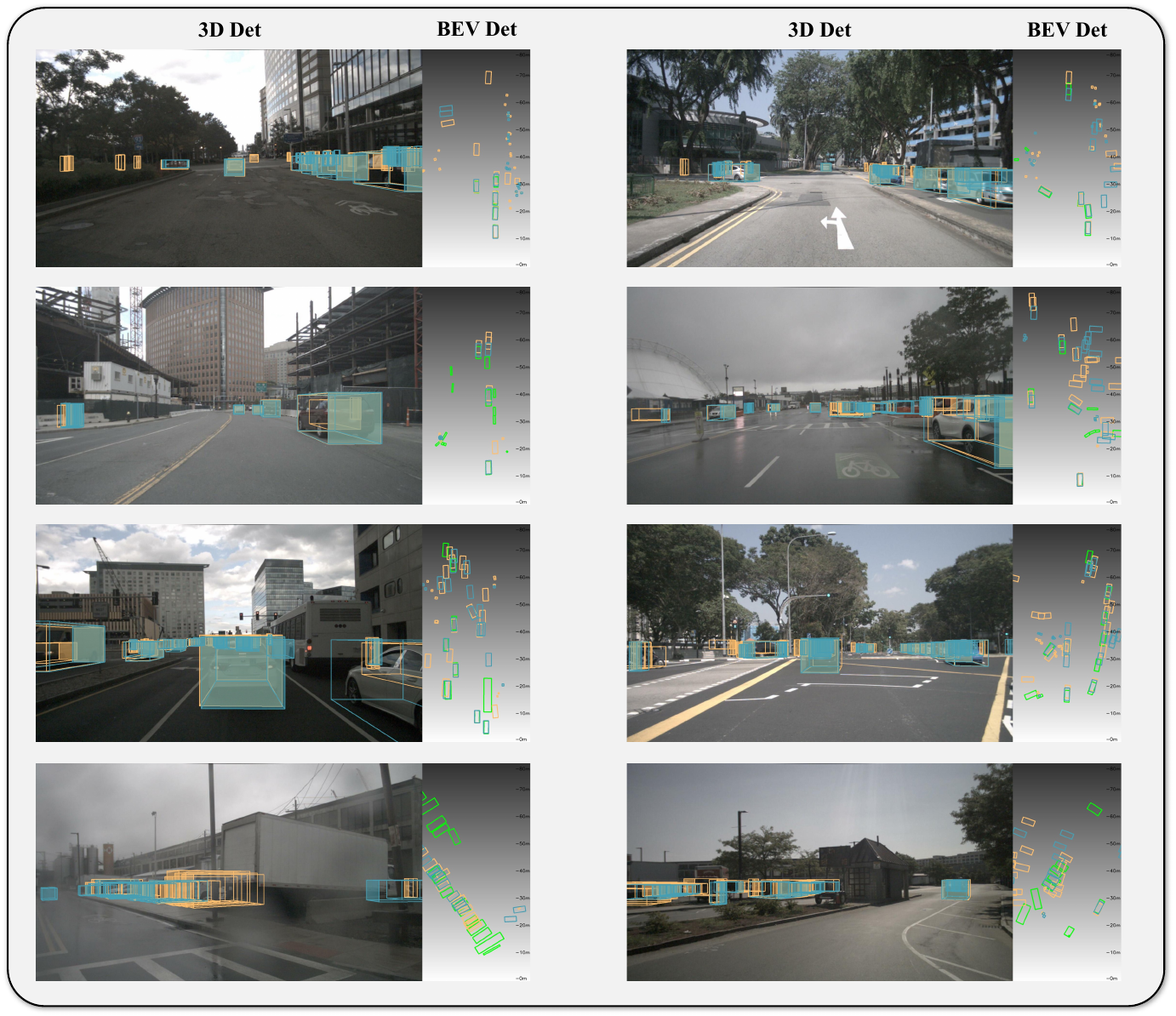}
    \caption{\textbf{\nuscenes \frontal Visualization.} \methodName demonstrates superior performance in both \threeD and BEV detection over the baseline~\cite{zhang2023monodetr}. By predicting \threeD attributes conditionally to mitigate instability and inaccuracy arising from their inter-correlations, \methodName predicts more accurate \threeD angle and depth, demonstrating effectiveness of \methodName. [Key: \textcolor{monocop}{\methodName}, \textcolor{baseline}{Baseline}, \textcolor{green}{Ground Truth}]}
    \vspace{-4mm}
    \label{nuscene-more}
\end{figure*}

We evaluate our \methodName on three well-known datasets: \kitti, \waymo, and \nuscenes. \methodName achieves \sota performance across these datasets. We finally visualize the detection results on these three datasets. \cref{kitti-more} presents the \threeD and BEV detection results. By predicting \threeD attributes conditionally to mitigate the instability and inaccuracy arising from their inter-correlation, \methodName improves detection accuracy, particularly for farther away objects, consistent with the results in Fig.~\ref{fig:error}c.  Similarly, \cref{waymo-more} demonstrates that, despite the larger variation in \threeD size in \waymo compared to \kitti, \methodName reliably predicts more accurate \threeD size and depth for large objects. Moreover, as shown in \cref{nuscene-more}, our method also delivers more precise angle and depth estimations.

\section{Ablations}
\label{appx-more-ablations}

\subsection{Things We Tried That Did Not Make it into the Main Algorithm }
\begin{figure}[t]
    \centering
    \includegraphics[width= 0.9 \linewidth]{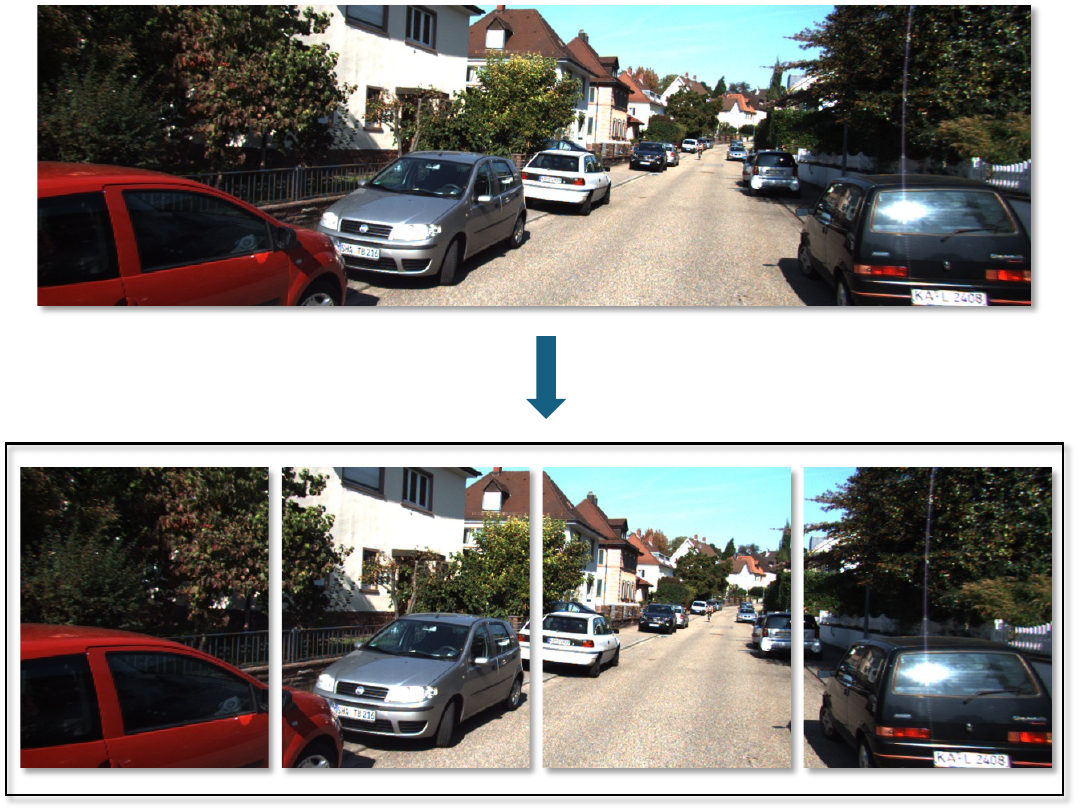}
    \caption{\textbf{Image Splitting.} The high-resolution original image is divided horizontally into four sub-images.}
    \label{kitti-split}
\end{figure}

\begin{itemize} 
    \item \textbf{Using DINOv$2$ \cite{oquabdinov2} as a Backbone.} We attempted to replace the conventional ResNet backbone in \monoThreeD with DINOv$2$, a powerful vision foundation model known for its depth perception capabilities. We experimented with both freezing and fine-tuning DINOv$2$ but found no performance improvement. We attribute this to (1) the relatively small scale of the \monoThreeD dataset, which may not fully leverage DINOv$2$’s capacity, and (2) the substantial domain gap between DINOv$2$’s pre-training data and \monoThreeD. 
    \item \textbf{Splitting Images into Sub-Images.} We also explored splitting the original image into four sub-images (shown in \ref{kitti-split}) and extracting features from each separately, motivated by the high resolution of the input images (e.g., $1280 \times 340$ in \kitti). Unfortunately, this approach led to inferior performance compared to using the entire image at once. 
    \item \textbf{Relation Encoding.} We additionally experimented with modeling pairwise relations between queries by incorporating their relative spatial positions. The goal is to enhance the detector’s geometric reasoning by providing explicit relational cues. However, we did not observe performance gains from this design.
    
\end{itemize}

\subsection{Different Backbones}

\begin{table}[t]
    \centering
    \begin{tabular}{l|c|ccc}
    \toprule
         \multirow{2}{*}{\makecell{Methods}} & \multirow{2}{*}{\makecell{Image \\ Backbone}} & \multicolumn{3}{c}{\apThreeD, $0.7$} \\
        & & Easy & Mod. & Hard \\
        \midrule
        \monodgp~\cite{pu2024monodgp} & ResNet-18 & 25.32 & 19.62 & 16.89 \\
        \rowcolor{lightgray} \methodName  & ResNet-18 & \textbf{27.78} & \textbf{21.03} & \textbf{17.98} \\
        \midrule
        \monodgp~\cite{pu2024monodgp} & ResNet-34 & 27.96 & 20.13 & 17.19 \\
        \rowcolor{lightgray} \methodName  & ResNet-34 & \textbf{28.32} & \textbf{22.32} & \textbf{19.23} \\
        \midrule
        \monodgp~\cite{pu2024monodgp} & ResNet-50 & 29.41 & 21.12 & 18.11 \\
        \rowcolor{lightgray} \methodName  & ResNet-50 & \cellcolor{lightgray}\textbf{32.06} & \cellcolor{lightgray}\textbf{23.98} & \cellcolor{lightgray}\textbf{20.64} \\
        \midrule
        \monodgp~\cite{pu2024monodgp} & ResNet-101 & 27.02 & 19.92 & 17.07 \\
        \rowcolor{lightgray} \methodName  & ResNet-101 & \textbf{30.14} & \textbf{21.75} & \textbf{18.56} \\
    \bottomrule
    \end{tabular}
    \caption{\textbf{Performance on Image backbone.} \methodName consistently outperforms MonoDGP~\cite{pu2024monodgp} across all backbones.}
    \label{sec:experiment-ablation:backbones}
\end{table}

In \cref{sec:experiment-ablation:backbones}, we evaluate \methodName with different image backbones on the \kitti \val split and observe that it consistently surpasses \monodgp~\cite{pu2024monodgp} under all configurations. Among the evaluated backbones, ResNet-50 yields the strongest overall detection performance.

\begin{table}[t]
    \centering
    \resizebox{0.35\textwidth}{!}{
    \begin{tabular}{c|ccc}
    \toprule
         \multirow{2}{*}{\makecell{AttributeNet}} & \multicolumn{3}{c}{\apThreeD, $0.7$} \\
        & Easy & Mod. & Hard \\
        \midrule
        LR         & 29.72 & 22.68 & 19.59 \\
        LR + ReLU  & 30.62 & 22.94 & 19.91 \\
        \rowcolor{lightgray}  2LR + ReLU  & \cellcolor{lightgray}\textbf{32.06} & \cellcolor{lightgray}\textbf{23.98} & \cellcolor{lightgray}\textbf{20.64} \\
        3LR + ReLU & 30.82 & 23.19 & 19.82 \\
    
    \bottomrule
    \end{tabular}}
    \caption{\textbf{Performance comparison of different AttributeNet (AN) designs.} We examine a single linear layer, our default two-layer MLP, and deeper variants. The two-layer configuration consistently delivers the best results, demonstrating its effectiveness in balancing representational capacity and computational cost.}
    \label{sec:experiment-an}
    \vspace{-2mm}
\end{table}

\subsection{Design of AttributeNet}
\methodName leverages an AttributeNet (AN) to capture attribute-specific features. Inspired by the MLP-based projector in vision-language models~\cite{liu2023llava}, we initially design AN as two linear layers with ReLU activation. This simple, two-layer structure strikes a balance between representational capacity and computational cost, allowing the model to effectively learn attribute representations without excessive overfitting. We then explore alternative AN configurations, such as a single linear layer or deeper MLP variants with additional linear layers and ReLU activations. As shown in \cref{sec:experiment-an}, however, our original two-layer configuration consistently yields the strongest overall performance, underscoring its efficacy in learning robust and discriminative attribute-specific features.

\begin{table}[t]
    \centering
    \resizebox{0.35\textwidth}{!}{
    \begin{tabular}{c|ccc}
    \toprule
         \multirow{2}{*}{\makecell{Number of \\ chain}} & \multicolumn{3}{c}{\apThreeD, $0.7$} \\
        & Easy & Mod. & Hard \\
        \midrule
    \rowcolor{lightgray}  One chain  & \cellcolor{lightgray}\textbf{32.06} & \cellcolor{lightgray}\textbf{23.98} & \cellcolor{lightgray}\textbf{20.64} \\
    Two chains   & 30.26 & 23.35 & 20.10 \\
    Three chains & 30.67 & 23.11 & 19.90 \\
    
    \bottomrule
    \end{tabular}}
\caption{\textbf{Performance of \methodName when varying the number of appended chains}. While adding extra chains can lead to marginal gains, the results demonstrate diminishing returns beyond the first chain, indicating that a single chain already captures most of the essential inter-attribute correlations.}
    \label{sec:experiment-number-of-chain}
    \vspace{-4mm}
\end{table}

\subsection{Number of Chain}
\methodName leverages a Chain-of-Prediction (CoP), which sequentially and conditionally predicts attributes by \textbf{learning}, \textbf{propagating}, and \textbf{aggregating} attribute-specific features along the chain. This design helps mitigate inaccuracies and instabilities arising from inter-correlations among 3D attributes.
In this subsection, we investigate how varying the number of chains in \methodName affects performance. First, we incorporate one additional chain and average the outputs across both chains. Next, we add two additional chains and average the outputs of all three. Our experimental findings (see \ref{sec:experiment-number-of-chain}) indicate that, although appending extra chains slightly increases computational complexity, it does not consistently yield notable performance gains. One plausible explanation is that the network may have already learned sufficient inter-attribute correlations from a single chain, causing further additions to become redundant. Another possible reason is that the increased complexity could introduce noise into the learning process, offsetting any potential benefits from extra chains. As a result, increasing the number of chains beyond one does not appear to offer further improvements in predictive accuracy.
\begin{table}[t]
    \centering
    \resizebox{0.49\textwidth}{!}{
    \begin{tabular}{l|ccc|ccc}
        \toprule
        \multirow{2}{*}{Method} & \multicolumn{3}{c|}{Ped \apThreeD{} \% (\uparrowRHDSmall{})} & 
        \multicolumn{3}{c}{Cyc \apThreeD{} \% (\uparrowRHDSmall{})} \\ 
        & Easy & Mod. & Hard & Easy & Mod. & Hard \\ 
        \midrule
        \monoflex~\cite{zhang2021objects}   & 11.89 & 8.16 & 6.81 & 3.39 & 2.10 & 1.67 \\
        \gupNet~\cite{lu2021geometry}    & 14.72 & 9.53 & 7.87 & 4.18 & 2.65 & 2.09 \\
        \deviant   ~\cite{kumar2022deviant}    & 15.04 & 9.89 & 8.38 & 5.28 & 2.82 & 2.65 \\
        \monocon{}~\cite{liu2022monocon}   & 13.10 & 8.41 & 6.94 & 2.80 & 1.92 & 1.55 \\
        \monouni~\cite{jia2023monouni}    & \textbf{15.78} & \textbf{10.34} & \textbf{8.74} & \underline{7.34} & \underline{4.28} & \underline{3.78} \\
        \monodgp~\cite{pu2024monodgp}    & 15.04 & 9.89 & 8.38 & 5.28 & 2.82 & 2.65 \\
        \midrule
        \textbf{\methodName  (Ours)} & 
        \underline{15.61} & \underline{10.33} & \underline{8.53} & 
        \textbf{8.89} & \textbf{5.08} & \textbf{5.25} \\
        \bottomrule
    \end{tabular}}
    \caption{
        \textbf{\kitti \test Results for Pedestrians and Cyclists} at \iouThreeD     $\geq 0.5$. \methodName  achieves \sota~performance across most metrics among image-only methods. [Key: \textbf{First}, \underline{Second}]}
    \label{exp:test-ped-cyclist}  
\end{table}

\section{\kitti Results}
\cref{exp:test-ped-cyclist} presents the image-only \threeD detection results on the \kitti \test for the Cyclist and Pedestrian categories. 
\methodName  achieves \sota performance across all metrics for the challenging Cyclist category and attains second-best results in the Moderate and Hard settings for the Pedestrian.

\section{Limitations.}

While \methodName models the interdependencies among 3D attributes and improves both accuracy and stability, it does not explicitly consider the influence of camera parameters. 
For instance, variations in camera focal length can introduce a zoom effect that alters the apparent scale of objects and may confuse the detector. 
Developing methods that remain robust under varying camera intrinsics is an important direction for future work. 
Moreover, recent advances in multimodal learning and recognition systems have demonstrated strong capabilities in visual reasoning and multimodal fusion~\cite{zhu2026fusionagent,chen2023atm,zhu2025quality,zhu2026can,su2026localscore,guo2026holistic,chen2025unlearning, chen2025safety}. 
However, how to effectively leverage these models for \monoThreeD remains largely unexplored.

\end{document}